\documentclass[10pt,twocolumn,letterpaper]{article}
\usepackage[pagenumbers]{cvpr}
\usepackage[T1]{fontenc}
\usepackage{amsmath,amssymb,bm}
\usepackage{booktabs,graphicx,xcolor,colortbl}
\usepackage{algorithm,algpseudocode}
\usepackage{cuted,currfile,caption}
\usepackage{microtype}
\definecolor{PaperPurple}{HTML}{8156B5}
\newcommand{\ShadeBest}{\cellcolor{PaperPurple!45}}
\newcommand{\ShadeSecond}{\cellcolor{PaperPurple!33}}
\newcommand{\ShadeThird}{\cellcolor{PaperPurple!23}}
\newcommand{\ShadeFourth}{\cellcolor{PaperPurple!14}}
\newcommand{\ShadeFifth}{\cellcolor{PaperPurple!6}}
\definecolor{cvprblue}{rgb}{0.21,0.49,0.74}
\usepackage[pagebackref,breaklinks,colorlinks,allcolors=cvprblue]{hyperref}
\newcommand{\E}{\mathbb E}
\newcommand{\R}{\mathbb R}
\newcommand{\ours}{LG-DKGP}
\newcommand{\oursname}{\textbf{LG-DKGP (Ours)}}

\DeclareMathOperator{\rvec}{rvec}

\DeclareMathOperator{\KL}{KL}
\DeclareMathOperator{\PE}{PE}
\DeclareMathOperator{\softmax}{softmax}

\newcommand{\PerformanceRows}{%
LeWM & \ShadeFourth $93.60\pm1.67$ & \ShadeFifth $63.60\pm9.74$ & \ShadeFourth $55.31\pm5.11$ & \ShadeFourth $69.88\pm8.48$ & \ShadeFourth $10.91\pm2.53$ & \ShadeFifth 58.66 \\
iCEM & \ShadeSecond $96.40\pm2.61$ & \ShadeFourth $69.60\pm9.10$ & \ShadeFifth $54.42\pm4.57$ & \ShadeThird $72.21\pm8.56$ & \ShadeFifth $4.03\pm3.83$ & \ShadeFourth 59.33 \\
INTACT & \ShadeFifth $64.40\pm5.55$ & \ShadeSecond $92.00\pm3.74$ & \ShadeSecond $80.97\pm3.95$ & \ShadeBest $\mathbf{77.53\pm4.24}$ & \ShadeSecond $40.26\pm8.85$ & \ShadeSecond 71.03 \\
PRISM & \ShadeThird $94.00\pm2.45$ & \ShadeThird $87.60\pm4.98$ & \ShadeThird $56.78\pm4.07$ & \ShadeFifth $56.91\pm6.10$ & \ShadeThird $13.01\pm5.41$ & \ShadeThird 61.66 \\
\oursname & \ShadeBest $\mathbf{97.60\pm0.89}$ & \ShadeBest $\mathbf{98.80\pm1.79}$ & \ShadeBest $\mathbf{83.32\pm2.05}$ & \ShadeSecond $76.96\pm5.31$ & \ShadeBest $\mathbf{78.66\pm4.94}$ & \ShadeBest \textbf{87.07} \\
}

\newcommand{\ContinuityRows}{%
PushT & 1842 & 0.354 & 0.815 & [0.805, 0.825] \\
Cube & 973 & 0.360 & 0.866 & [0.859, 0.873] \\
Caddy & 1000 & 0.267 & 0.855 & [0.850, 0.859] \\
Scene2 & 985 & 0.299 & 0.746 & [0.742, 0.750] \\
RoboTwin & 1000 & 0.020 & 0.514 & [0.502, 0.526] \\
}

\def\confName{CVPR}
\def\confYear{2027}
\title{Rethinking Actions: Continuity-Aware Priors for World-Model Planning}

\newcommand{\InitializationRows}{%
PushT & 97.60 & 97.60 & 97.60 & 97.60 \\
Cube & 99.20 & 98.80 & 98.80 & 98.80 \\
Caddy & 83.34 & 76.97 & 76.98 & 83.32 \\
Scene2 & 75.56 & 69.44 & 72.98 & 76.96 \\
RoboTwin & 18.88 & 72.86 & 77.81 & 78.66 \\
\midrule
Macro & 74.92 & 83.14 & 84.84 & 87.07 \\
}

\newcommand{\PairedStructureRows}{%
PushT & 209/250 & 83.6 & $0.0846$ & $[0.0716,0.0976]$ \\
Cube & 225/250 & 90.0 & $0.2137$ & $[0.1958,0.2314]$ \\
Caddy & 221/222 & 99.5 & $0.3156$ & $[0.2985,0.3337]$ \\
Scene2 & 222/226 & 98.2 & $0.1480$ & $[0.1359,0.1601]$ \\
RoboTwin & 219/219 & 100.0 & $0.5178$ & $[0.4966,0.5350]$ \\
}

\usepackage{stfloats}
\author{
Shaobo Yang$^{1,*}$\quad
Zichen Ai$^{1,*}$\quad
Ziyang Ma$^{1,*}$\quad
Qian Yang$^{2,*}$\quad
Bin Fang$^{1,\dagger}$\\[3pt]
$^1$Beijing University of Posts and Telecommunications\\
$^2$Alibaba Group\\[3pt]
{\small\ttfamily\{yangshaobo,aizichen,maziyang,fangbin1120\}@bupt.edu.cn}\\
{\small\ttfamily chashi.yq@antgroup.com}\\[3pt]
{\small $^*$Equal contribution.\quad $^\dagger$Corresponding author.}\\[3pt]
{\small Code: \url{https://github.com/oliverstack-byte/LG-DKGP}}
}
\hypersetup{
  pdftitle={Rethinking Actions: Continuity-Aware Priors for World-Model Planning},
  pdfauthor={Shaobo Yang, Zichen Ai, Ziyang Ma, Qian Yang, Bin Fang},
  pdfsubject={Research preprint}
}

\begin{document}
\raggedbottom
\maketitle
\begin{strip}
\centering
\includegraphics[width=0.85\textwidth]{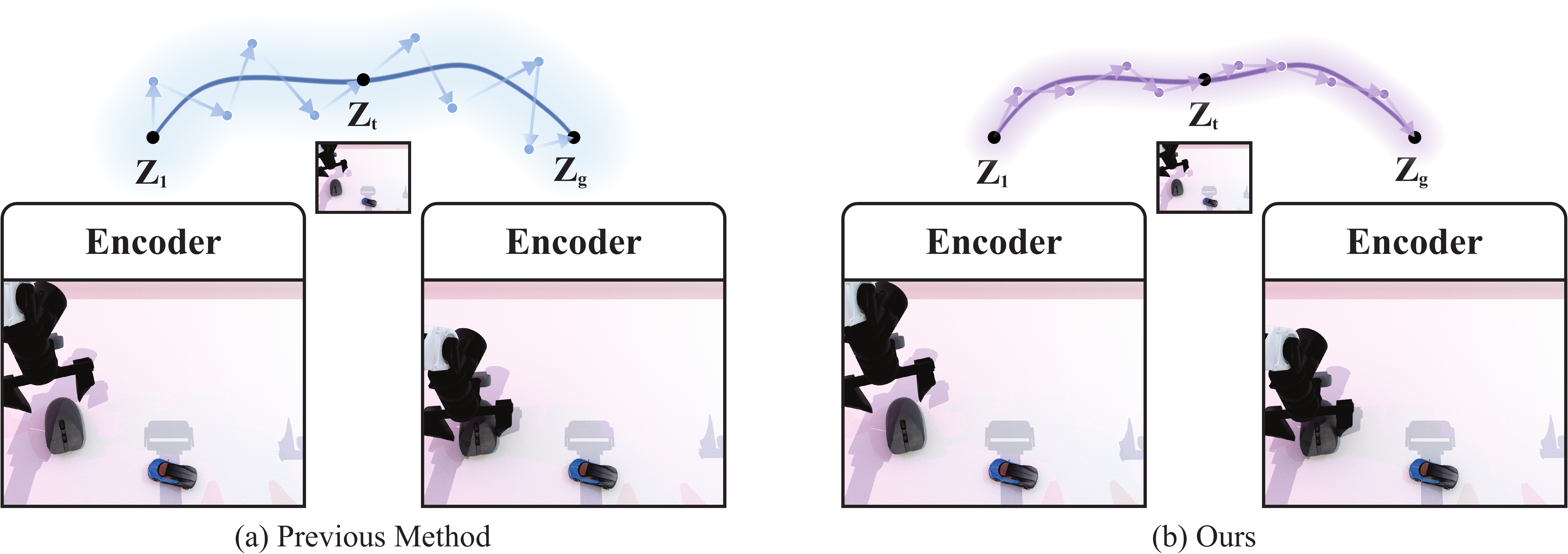}
\captionsetup{hypcap=false}
\captionof{figure}{\textbf{Rethinking actions as demonstrated continuations.} The
schematic contrasts unconstrained proposals with an expert-informed action
prior. A frozen visual model evaluates candidate continuations toward a goal;
our contribution shapes the actions proposed to it.}
\label{fig:overview}
\captionsetup{hypcap=true}
\end{strip}

\begin{abstract}
Actions in expert demonstrations exhibit temporal structure. An
ordered-versus-shuffled analysis across five environments confirms this
dependence. We ask whether action proposals that better reflect expert
continuity can improve visual world-model planning. Rather than imposing
generic smoothness, we learn continuations from the temporal structure of
demonstrated actions. Our locality-gated
deep-kernel Gaussian-process prior, LG-DKGP, combines recent and longer-range
executed action history, predicts an action continuation, and learns a temporal
Gaussian-process (GP) residual. A training-derived exploration scale initializes search around
the corrected mean; a frozen world model evaluates progress toward the goal.
Training first fits the neural mean, then jointly optimizes the prior with
the GP residual using expert future actions.
The resulting planner achieves 87.07\% macro success across five simulated
tasks spanning native and fragment-level goal attainment, compared with
71.03\% for INTACT, 61.66\% for PRISM, and 58.66\% for LeWM.
Component ablations and expert-relative action-structure comparisons
support expert-informed continuations and reveal task-dependent effects
of visual conditioning, history length, and planning horizon.
\end{abstract}

\section{Introduction}
\label{sec:intro}
Expert behavior is organized in time: reaching commands follow ongoing motion,
and manipulation commands depend on contact phase. Treating future actions
as unconstrained vectors overlooks this temporal structure.

Across held-out episodes from five environments, preserving expert action
order reduces adjacent standardized changes relative to within-episode
permutations. We ask:
\emph{Can proposals that better reflect the continuity structure of expert
action sequences improve goal-directed world-model planning?}

Visual world models provide a natural evaluation interface: predict the outcomes of candidate
actions and compare the predicted latent state with a visual goal.
Joint-embedding predictive architectures learn image and video features through
embedding-space prediction~\cite{assran2023ijepa,bardes2024vjepa}.
LeWorldModel (LeWM) learns such action-conditioned predictors from pixels~\cite{maes2026lewm}.
PRISM guides imagination with a learned action prior~\cite{wang2026prism},
whereas INTACT learns an intent-to-action mapping for direct
execution~\cite{sun2026intact}.

We introduce \ours{}, a locality-gated deep-kernel Gaussian-process action
prior. A learned gate combines local and global attention over executed
actions. A neural mean and temporal GP residual predict continuations;
expert training increments calibrate exploration around their combined mean.
The frozen world model evaluates goal attainment (Fig.~\ref{fig:overview}).

The target is continuity \emph{exhibited by experts}: constant commands can
have low variation yet be inappropriate. We assess both control success and
discrepancy from expert transition structure.

Our contributions are:
\begin{itemize}
\item A held-out action-order study that motivates expert-relative continuity as a design principle for world-model planning.
\item A locality-gated neural/GP action prior that combines local and global action-history information with training-derived search initialization.
\item A five-task evaluation of control performance, component ablations, history/horizon sensitivity, and expert-relative transition structure.
\end{itemize}

\section{Related work}
\label{sec:related}
\paragraph{Visual world models.}
Predictive embeddings~\cite{assran2023ijepa,bardes2024vjepa,li2026salt},
visual dynamics~\cite{oquab2024dinov2,zhou2025dinowm,maes2026lewm},
latent control~\cite{hafner2019planet,hafner2025dreamerv3,hansen2024tdmpc2},
decoder-free and offline planning~\cite{morihira2026r2dreamer,sobal2025pldm},
and geometric/closed-loop models~\cite{zhang2026geoworld,zhang2026worldinworld}
advance prediction and representation. LeWM's latent goal objective evaluates
outcomes but does not explicitly encourage expert-like action transitions.
\ours{} addresses this action-side gap by learning demonstrated
continuations while retaining the frozen world model for goal evaluation.

\paragraph{Action representations and planning.}
Latent actions~\cite{bruce2024genie,garrido2026latentaction}, trajectory
diffusion~\cite{janner2022diffuser}, action chunking~\cite{zhao2023act}, and
diffusion policies~\cite{chi2023diffusionpolicy} learn structured sequences.
\ours{} explicitly connects proposals to executed actions through a gate
that dynamically combines recent motion and longer-range action context.
FLIP couples flow proposals with visual prediction~\cite{gao2025flip};
parallel gradient planning~\cite{psenka2026parallel} and correlated iCEM
sampling~\cite{pinneri2020icem} improve optimization. iCEM prescribes noise
correlation; \ours{} learns continuation means and calibrates exploration
from expert increments. The target is demonstrated temporal structure,
which small action changes alone cannot establish.

PRISM conditions its prior on current/goal embeddings~\cite{wang2026prism},
without explicit executed-action inputs. \ours{} adds gated local/global
action history and a neural/GP continuation mean.
INTACT learns an intent-to-action
interface~\cite{sun2026intact}; its direct mode omits candidate evaluation,
which optional CEM provides. \ours{} retains goal-directed search around
expert-informed continuations. Table~\ref{tab:main} compares performance;
Fig.~\ref{fig:search} evaluates the prior components and subsequent search.

\paragraph{Gaussian-process priors.}
Deep kernel learning combines neural features with GP regression~\cite{wilson2015deep}; adaptation to low intrinsic
dimension~\cite{yang2016manifold} and sparse variational
inference~\cite{titsias2009variational,hensman2013bigdata} motivate our compact
residual model. Our contribution integrates these tools with gated action
history and expert future-action supervision: a neural mean captures nonlinear
continuations, and the GP learns context-dependent residual dependence across
forecast positions. Continuous covariance provides process
regularity~\cite{rasmussen2006gp}; expert increments calibrate exploration.
We assess both task success and expert-relative transition structure.

\section{Problem formulation and motivation}
\label{sec:formulation}
Given an observation $o_t$ and visual goal $o_g$, a frozen encoder $E_\omega$
produces latents $\mathbf z_t,\mathbf z_g\in\R^{d_z}$. A frozen predictor
$F_\omega$ predicts outcomes of candidate actions. We seek a chunk of $H$ actions,
execute its first $P$ actions, then observe again and replan. Bold lowercase and
uppercase symbols denote vectors and matrices; $\ell\in\{1,\ldots,H\}$ is
forecast lead, $j$ is action coordinate, and $r$ is search iteration.

We standardize actions using training-set means and standard deviations,
$\widetilde{\mathbf a}_t=(\mathbf a_t-\boldsymbol\mu_a)
\oslash\boldsymbol\sigma_a$, where $\oslash$ denotes element-wise division.
The candidate sequence
$\widetilde{\mathbf a}_{t:t+H-1}\in\R^{H\times d}$ stacks future standardized
actions as rows. The executed action history
$\widetilde{\mathbf a}_{t-L:t-1}\in\R^{L\times d}$ contains only commands
before $t$, with left padding and a validity mask $\boldsymbol\chi_t$.
The prior conditions on the information tuple
\begin{equation}
 \mathcal C_t=(\bar{\mathbf z}_t,\bar{\mathbf z}_g,
              \widetilde{\mathbf a}_{t-L:t-1},\boldsymbol\chi_t).
\end{equation}
Here the prior uses feature-wise standardized latents,
\[
 \bar{\mathbf z}_v=(\mathbf z_v-\boldsymbol\mu_z)
                  \oslash\boldsymbol\sigma_z,\quad v\in\{t,g\},
\]
where training-set statistics $\boldsymbol\mu_z,\boldsymbol\sigma_z$ also
standardize current/goal features at inference. World-model prediction and
scoring retain the original latents $\mathbf z_t,\mathbf z_g$.
Expert future actions supply supervision and offline references.

\paragraph{What continuity means here.}
Our empirical motivation is local temporal dependence in expert commands.
For continuous coordinates, we compute the RMS difference between consecutive standardized actions,
averaged within each episode, $\bar\delta_1$. We then permute action vectors
within each episode, retaining their values but changing their order.
Table~\ref{tab:continuity} reports shuffled-minus-ordered differences
$\Delta_{\mathrm{ord}}$. All five effects are positive, with episode-bootstrap
intervals excluding zero. Gripper transitions are excluded from this particular
statistic and analyzed separately.

\begin{table}[t]
\centering
\caption{\textbf{Expert actions have local temporal structure.}
Held-out successful episodes; 95\% intervals use episode bootstrap.
$\Delta_{\mathrm{ord}}$ is shuffled minus ordered adjacent difference.}
\label{tab:continuity}
\setlength{\tabcolsep}{2.5pt}\footnotesize
\begin{tabular*}{\columnwidth}{@{\extracolsep{\fill}}lrrrr@{}}
\toprule
Task & Episodes & $\bar\delta_1$ & $\Delta_{\mathrm{ord}}$ & 95\% interval\\
\midrule
\ContinuityRows
\bottomrule
\end{tabular*}
\end{table}

\begin{figure}[t]
\centering
\includegraphics[width=0.85\columnwidth]{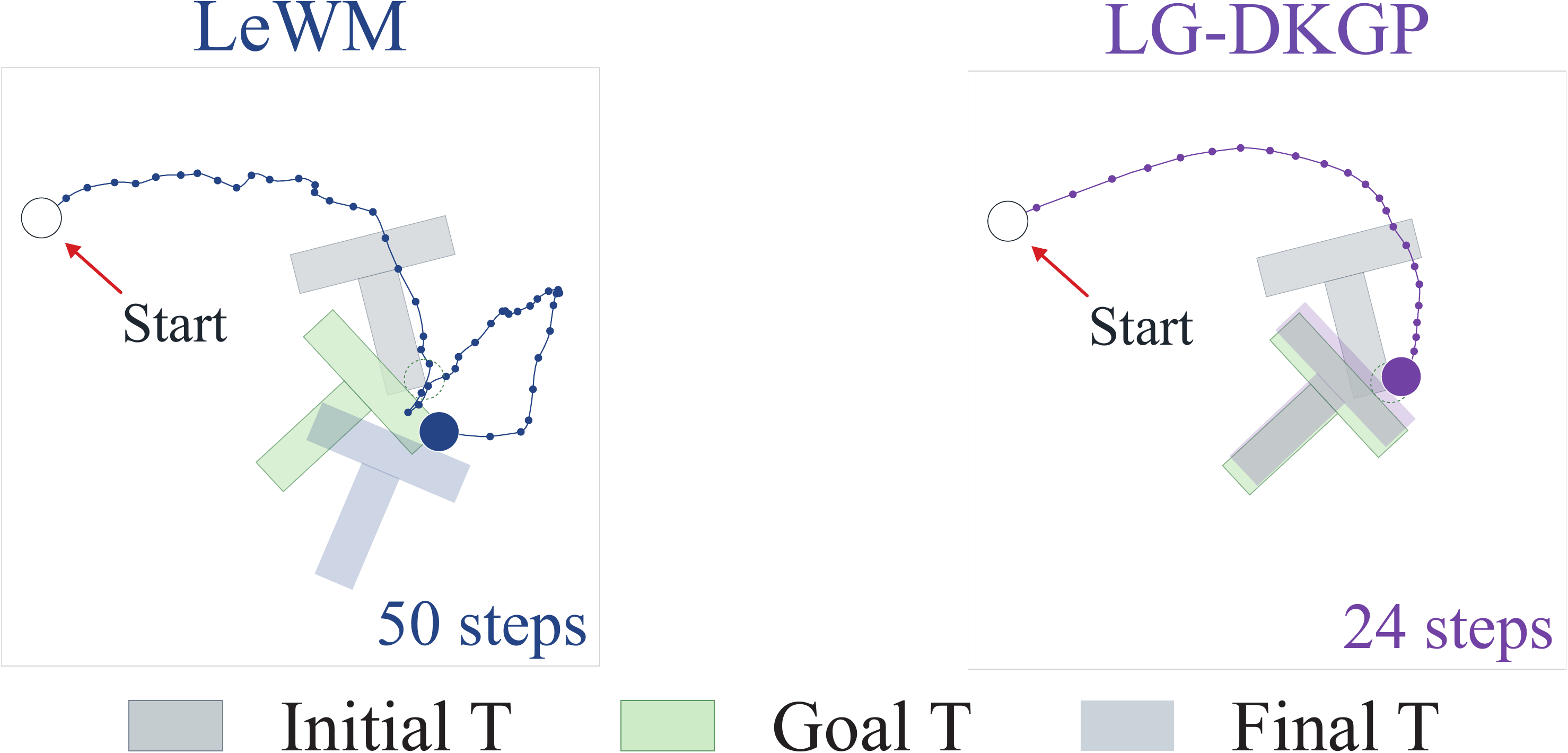}
\caption{\textbf{Matched PushT executions.} Both methods share the start and
goal. LG-DKGP succeeds at step 24; LeWM fails within 50 steps. Lines follow
the actual circular pusher, dots mark control-step positions, and T shapes
show initial, goal, and final object poses.}
\label{fig:pusherpaths}
\end{figure}

Figure~\ref{fig:pusherpaths} shows LeWM turning near the object and leaving
it misaligned; LG-DKGP reaches the shared goal through a coherent push.

We encode this local dependence using a mean-square continuous conditional
action process: for a noise-free action process $\mathbf g(\ell;\mathcal C)$,
$\E\|\mathbf g(\ell;\mathcal C)-\mathbf g(\ell';\mathcal C)\|^2\to0$ as
$\ell'\to\ell$ for continuous prediction times. Continuous means and covariance
kernels supply this regularity. Agreement with the expert continuation is
measured by transition-feature discrepancy and variation-profile error.

\section{Continuity-aware action priors}
\label{sec:method}
\begin{figure*}[t]
\centering
\includegraphics[width=0.95\textwidth]{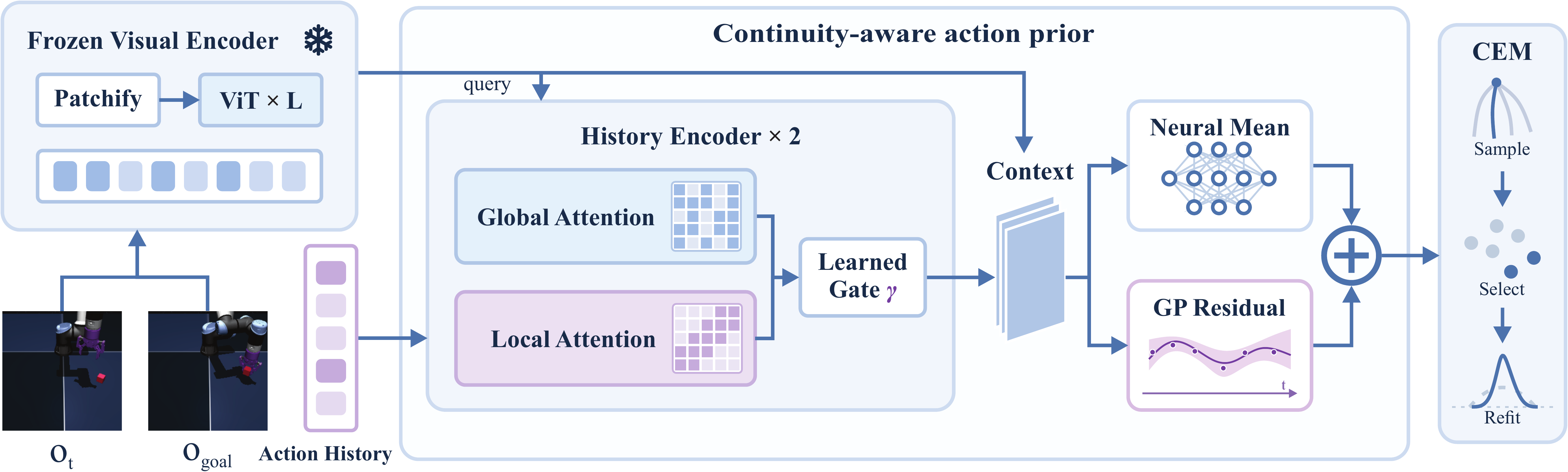}
\caption{\textbf{Continuity-aware action prior and planning.} Frozen visual
features query a two-layer history encoder. Learned gating combines global
and recency-biased local attention over executed actions. A neural mean and
GP residual form the continuation mean supplied to CEM, which explores
its neighborhood using a training-derived exploration scale.}
\label{fig:architecture}
\end{figure*}

Figure~\ref{fig:architecture} summarizes \ours. The action prior constructs a
plausible continuation; the world model evaluates its progress toward the goal.
Current and goal features query the executed-action history, whose gated
summary conditions the neural mean and GP residual. Their combined mean
centers the candidate search before frozen-model evaluation and execution.

\subsection{Locality-gated executed action history}
Recent actions reveal the current motion trend, whereas older commands can
help disambiguate its phase. We therefore
combine local and global attention through a learned, context-dependent gate.
Both branches use multi-head scaled dot-product attention~\cite{vaswani2017attention}.
Global attention ranges over all valid retained actions.

For the $i$th retained history action $\widetilde{\mathbf a}^{\rm hist}_i$, validity $\chi_i$, and age $\rho_i$, set
$\iota_i=\chi_i\chi_{i-1}$, with $\chi_0=0$. Its token is
\begin{equation}
 \mathbf e_i=\chi_i[\mathbf W_a\widetilde{\mathbf a}^{\rm hist}_i+
 \mathbf W_\Delta\Delta\widetilde{\mathbf a}^{\rm hist}_i+\mathbf w_\iota\iota_i+\PE(\rho_i)].
\end{equation}
Here $\Delta\widetilde{\mathbf a}^{\rm hist}_i=\widetilde{\mathbf a}^{\rm hist}_i-\widetilde{\mathbf a}^{\rm hist}_{i-1}$ only when both actions
are valid, and is zero otherwise. Current/goal features initialize the attention query as
$\mathbf q^{(0)}=\mathbf W_q[\bar{\mathbf z}_t;
\bar{\mathbf z}_g;\bar{\mathbf z}_g-\bar{\mathbf z}_t]$.
Semicolons indicate concatenation.

Let $\mathcal V=\{i:\chi_i=1\}$ and
$\zeta_i=\langle\mathbf W_Q\mathbf q,\mathbf W_K\mathbf e_i\rangle/\sqrt{d_k}$.
Each attention head computes
\begin{align}
 \alpha_i^{\rm glob}&=\softmax_{i\in\mathcal V}(\zeta_i),\nonumber\\
 \alpha_i^{\rm loc}&=\softmax_{i\in\mathcal V}
                (\zeta_i-\lambda(\mathbf q)\rho_i),
\end{align}
where $\lambda(\mathbf q)>0$ is learned via softplus. The local branch softly
favors recency while preserving access to older actions. Stack token rows
as $\mathbf E=[\mathbf e_1,\ldots,\mathbf e_L]^\top$, and let
$\boldsymbol\alpha^{b,(s)}\in\R^{1\times L}$ contain the attention weights
of branch $b\in\{\mathrm{loc},\mathrm{glob}\}$ and head $s$, with zero
weights at masked positions. Value projection and aggregation are
\begin{equation}
 \begin{aligned}
 \mathbf V^{(s)}&=\mathbf E(\mathbf W_V^{(s)})^\top,\\
 \mathbf h^{b}&=\left[\operatorname{Concat}_{s=1}^{N_h}
       (\boldsymbol\alpha^{b,(s)}\mathbf V^{(s)})\right]^\top.
 \end{aligned}
 \label{eq:historysummary}
\end{equation}
Here $N_h=4$; the head index is suppressed in the attention-weight equations
above. The product $\boldsymbol\alpha^{b,(s)}\mathbf V^{(s)}$ is the
attention-weighted value summary of one head. Thus $\mathbf h^{\rm loc}$
and $\mathbf h^{\rm glob}$ summarize the same executed action tokens using
different temporal weighting. We fuse
these summaries with a learned scalar gate:
\begin{align}
 \gamma&=\operatorname{sigmoid}
       (g_\theta[\mathbf q;\mathbf h^{\rm loc};\mathbf h^{\rm glob}]),\nonumber\\
 \mathbf h&=(1-\gamma)\mathbf h^{\rm loc}+\gamma\mathbf h^{\rm glob}.
 \label{eq:gate}
\end{align}
Each layer adds its fused history contribution $\mathbf h$ to the query.
For layer $k$, the update is
\begin{equation}
 \begin{aligned}
 \mathbf u^{(k)}&=\operatorname{LN}(\mathbf q^{(k)}+\mathbf h^{(k)}),\\
 \mathbf q^{(k+1)}&=\operatorname{LN}
   (\mathbf u^{(k)}+\operatorname{FFN}(\mathbf u^{(k)})).
 \end{aligned}
 \label{eq:historyupdate}
\end{equation}
Two layers with width 128 produce the final history-conditioned context
$\mathbf h_t=\mathbf q^{(2)}$; empty history instead uses a learned fallback
vector. History is capped at 32 actions. The neural context is
$\mathbf r_t=[\bar{\mathbf z}_t;\bar{\mathbf z}_g;\mathbf h_t]$.

\subsection{Neural mean with a temporal Gaussian-process (GP) residual}
The neural head maps $[\mathbf r_t;\PE(\ell)]$ to a mean action
$\mathbf m_\theta(\ell;\mathcal C_t)\in\R^d$. A GP residual captures
structure not modeled by the neural mean:
\begin{equation}
 \widetilde a_j(\ell;\mathcal C_t)=m_{\theta,j}(\ell;\mathcal C_t)
                     +f_j(\mathbf x_{t,\ell})+\epsilon_{\ell,j},
 \label{eq:residual}
\end{equation}
with Gaussian likelihood noise $\epsilon_{\ell,j}$ and independent
$f_j\sim\mathcal{GP}(0,k_j)$ across action coordinates. Learned compact
features form
$\mathbf x_{t,\ell}=[\boldsymbol\phi_\theta(\mathbf r_t,\ell);
\boldsymbol\psi_\theta(\mathbf r_t);\ell]$, with 32-dimensional feature maps.
The kernel combines query similarity with context-dependent temporal
correlation:
\begin{align}
 k_j(\mathbf x,\mathbf x')={}&\sigma_{f,j}^2
             k_{\rm RBF}(\boldsymbol\phi,\boldsymbol\phi')\nonumber\\
 &+\eta_j k_{\rm RBF}(\boldsymbol\psi,\boldsymbol\psi')
                   k_{\rm Mat\acute ern\,3/2}(\ell,\ell').
 \label{eq:kernel}
\end{align}
This construction models dependence across forecast positions for a fixed
context.

GP generalization can adapt to low intrinsic dimensional structure under
regularity assumptions~\cite{yang2016manifold}. This motivates compact
learned GP inputs~\cite{wilson2015deep}, since high-dimensional inputs can
impair kernel learning~\cite{delbridge2020projected}.

We use sparse variational inference with 128 inducing points per coordinate
and $q_j(\mathbf u_j)=\mathcal N(\mathbf m_{u,j},\mathbf S_{u,j})$
\cite{titsias2009variational}. With
$\mathbf C_{t,j}=\mathbf K_{X_tZ_j}\mathbf K_{Z_jZ_j}^{-1}$, the posterior mean
across leads is
\begin{align}
 \boldsymbol\mu_{t,j}&=\mathbf m_{\theta,t,j}
                         +\mathbf C_{t,j}\mathbf m_{u,j},\nonumber\\
 \mathbf M_t&=[\boldsymbol\mu_{t,1},\ldots,\boldsymbol\mu_{t,d}].
 \label{eq:posterior_mean}
\end{align}

\subsection{Learning demonstrated continuations}
\begin{figure}[t]
\centering
\includegraphics[width=\columnwidth]{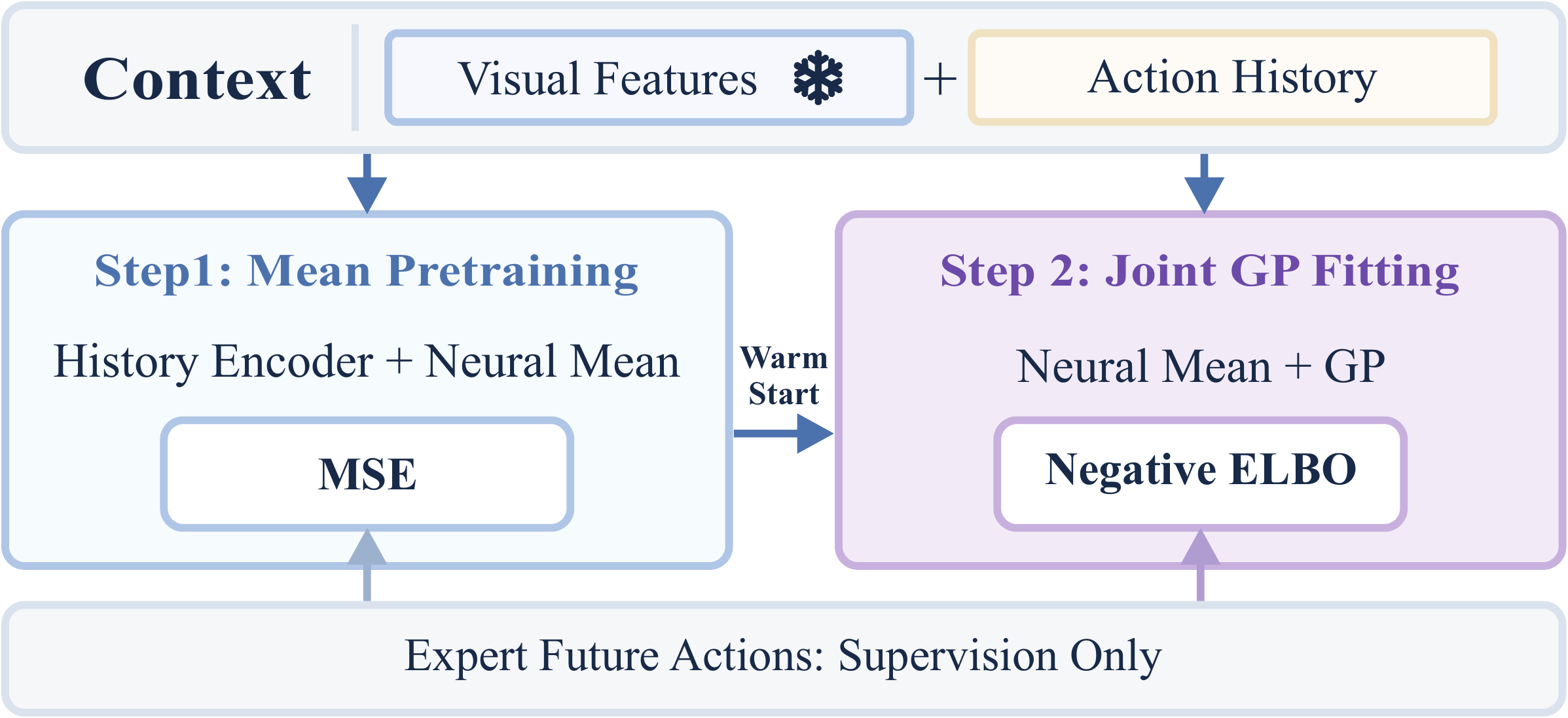}
\caption{\textbf{Two-stage prior learning.} A history encoder and neural mean
are pretrained using expert future-action targets, then warm-start joint
prior fitting with a GP residual. The second stage also updates the history
encoder. The visual encoder remains frozen. Future expert actions supervise
the losses and are unavailable as online conditioning inputs.}
\label{fig:training}
\end{figure}

Successful expert episodes supply training windows, with episode-disjoint
fitting and validation. Let $\widetilde a^{\rm exp}_{n\ell j}$ denote
coordinate $j$ of the standardized expert future action at forecast lead
$\ell$ in training window $n$. We first pretrain the history encoder and
neural mean:
\begin{equation}
 \mathcal L_{\rm MSE}=\frac{1}{N_{\rm tr}Hd}
       \sum_{n,\ell,j}[m_{\theta,j}(\ell;\mathcal C_n)-\widetilde a^{\rm exp}_{n\ell j}]^2.
\end{equation}
These weights initialize joint prior training (Fig.~\ref{fig:training}). We
maximize the variational evidence lower bound~\cite{titsias2009variational,hensman2013bigdata}
\begin{align}
 \mathcal L_{\rm ELBO}={}&\sum_{n,\ell,j}\E_{q_j(f_{n\ell j})}
    \log p(\widetilde a^{\rm exp}_{n\ell j}\mid m_{\theta,j}(\ell;\mathcal C_n)+f_{n\ell j})
    \nonumber\\
 &-\sum_j\KL(q_j(\mathbf u_j)\|p_j(\mathbf u_j)).
\end{align}
The history encoder, neural mean, learned kernel features, and GP parameters
remain trainable in the second stage. $E_\omega$ and $F_\omega$ stay frozen.
Validation MSE selects the pretraining checkpoint; joint predictive NLL selects
the GP-stage checkpoint.

\subsection{Expert-scaled search around the prior}
The mean supplies a continuation; search needs a scale for exploring its
neighborhood. For environment $e$ and training-episode adjacent pairs
$\mathcal P_e$, define
\begin{align}
 \widehat v_{\Delta,e,j}&=\frac{1}{|\mathcal P_e|}
     \sum_{(\tau,t)\in\mathcal P_e}
       (\widetilde a_{\tau,t+1,j}-\widetilde a_{\tau,t,j})^2,\nonumber\\
 \sigma_{\rm exp,e}^2&=\frac{1}{2d_e}\sum_{j=1}^{d_e}
                                  \widehat v_{\Delta,e,j}.
 \label{eq:sigma}
\end{align}
All coordinates, including grippers, enter this training statistic. The factor
$1/2$ matches the increment variance of independent Gaussian perturbations.
The scale is estimated from training actions and fixed across evaluation
seeds and planning calls. The initialization comparison in
Table~\ref{tab:initialization} informed this choice.

Let $\rvec(\widetilde{\mathbf a})=\operatorname{vec}(\widetilde{\mathbf a}^{\top})$ denote row-wise sequence vectorization.
Search starts from
\begin{equation}
 \rvec(\widetilde{\mathbf a}_{t:t+H-1})\sim\mathcal N
             (\rvec(\mathbf M_t),\sigma_{\rm exp,e}^2\mathbf I_{Hd}).
 \label{eq:proposal}
\end{equation}
CEM samples perturbations around this mean, scores candidates by
\begin{equation}
 J_t(\widetilde{\mathbf a})=\|F_\omega^{(H)}(\mathbf z_t,\widetilde{\mathbf a})-\mathbf z_g\|_2^2,
 \label{eq:cost}
\end{equation}
and refits a diagonal Gaussian to the elites. We use 300 candidates, 30
iterations, and 30 elites: 9,000 candidate scores per call. The returned chunk
is converted to environment units, and up to $P$ actions are executed. Fresh
observations and the commands actually executed rebuild the next context,
as in receding-horizon control~\cite{rawlings2026mpc}. 

\section{Experiments}
\label{sec:experiments}
\begin{table*}[t]
\centering
\caption{\textbf{Main performance across five environments (\%).} Task cells
are five-seed means $\pm$ sample SD; Macro is their unweighted mean.
Caddy, Scene2, and RoboTwin use the fragment-level criteria described in the
text. Bold numeric entries highlight the highest task mean. Five purple shades rank each column from higher (darker) to lower success (lighter).}
\label{tab:main}
\setlength{\tabcolsep}{5pt}\small
\begin{tabular*}{\textwidth}{@{\extracolsep{\fill}}lcccccc@{}}
\toprule
Method & PushT & Cube & Caddy & Scene2 & RoboTwin & Macro\\
\midrule
\PerformanceRows
\bottomrule
\end{tabular*}
\end{table*}

\subsection{Tasks, baselines, and evaluation}
\begin{figure*}[t]
\centering
\includegraphics[width=\textwidth]{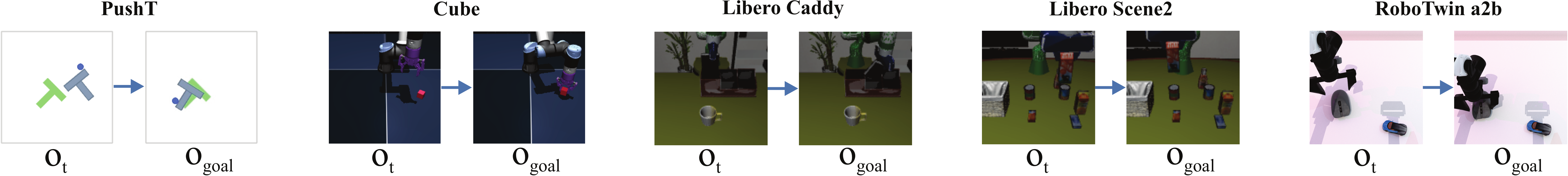}
\caption{\textbf{Five evaluation environments.} Current observations $\mathbf o_t$
and goal observations $\mathbf o_g$ illustrate the visual planning inputs for
PushT, Cube, LIBERO Caddy, LIBERO Scene2, and RoboTwin.}
\label{fig:environments}
\end{figure*}
Figure~\ref{fig:environments} shows example current and goal observations.
We evaluate PushT, OGBench Cube~\cite{park2025ogbench}, Caddy and Scene2 from LIBERO~\cite{liu2023libero}, and RoboTwin~\cite{chen2025robotwin}. PushT and Cube use
recorded native task-success predicates. Caddy and Scene2 measure attainment
of a goal sampled from a demonstration fragment. RoboTwin uses the same
local-goal-or-verified-native-terminal predicate across methods. Macro gives equal weight to the five task-level rates.

The main table uses evaluation seeds 101, 202, 303, 404, and 505. Each task
cell is the unweighted mean of seed success rates with sample standard
deviation (SD). For successes $S_{m,e,s}$ and eligible episodes $N_{m,e,s}$,
\begin{align}
 \bar p_{m,e}&=\frac{1}{5}\sum_s\frac{100S_{m,e,s}}{N_{m,e,s}},\nonumber\\
 p_m^{\rm macro}&=\frac{1}{5}\sum_e\bar p_{m,e}.
 \label{eq:success_aggregation}
\end{align}
We compare original LeWM planning, iCEM~\cite{pinneri2020icem},
PRISM's released action-head approach~\cite{wang2026prism}, and
INTACT, trained following its original task-specific one-epoch setting and
executed directly~\cite{sun2026intact}. The main setting uses $H=P=25$
and a 50-step execution budget.

\subsection{Main performance}
\begin{figure}[t]
\centering
\includegraphics[width=\columnwidth]{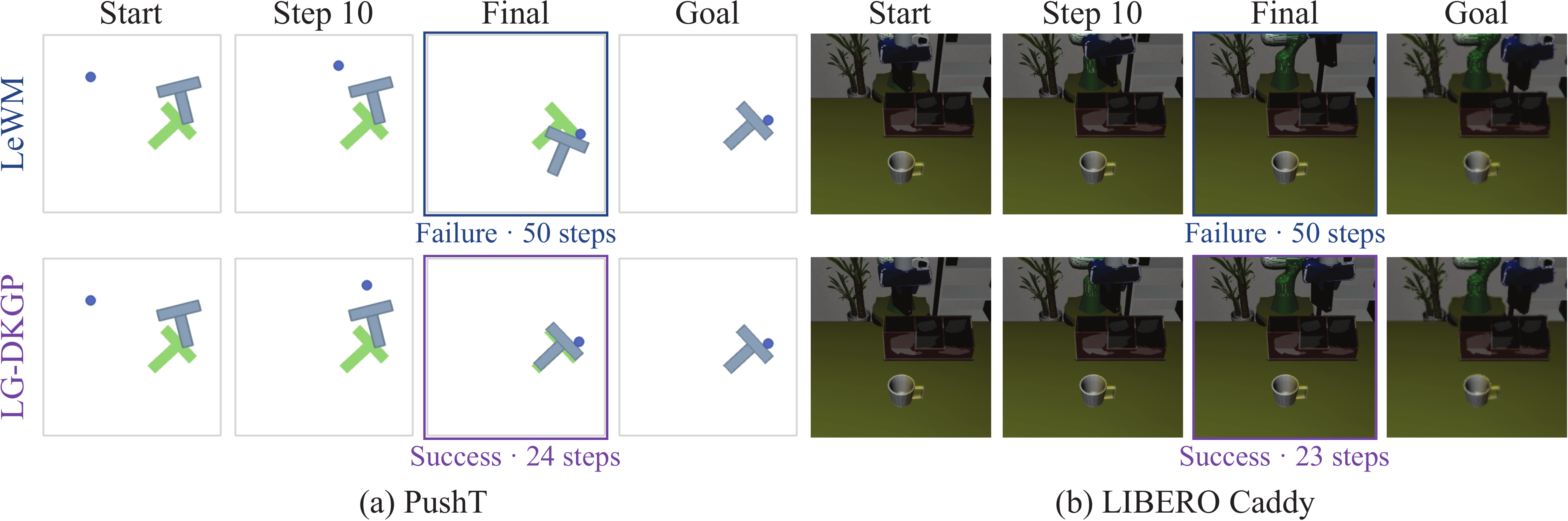}
\caption{\textbf{Paired execution examples.} PushT and Caddy use matched
starts and goals for LeWM and LG-DKGP. Columns show step 10 and the recorded
final state, with step counts indicated. LG-DKGP reaches the native PushT
criterion and the Caddy fragment goal; LeWM fails within 50 steps.}
\label{fig:pairedexecutions}
\end{figure}
LG-DKGP achieves 87.07\% macro success (Table~\ref{tab:main}), compared with
71.03\% for INTACT, 61.66\% for PRISM, and 58.66\% for LeWM. The respective
gaps are 16.04, 25.41, and 28.41 percentage points. Our method has the
highest mean in four environments. Scene2 is an exception: INTACT reaches
77.53\%, compared with 76.96\% for LG-DKGP. PushT and Cube approach ceiling
performance; the larger separations on Caddy and RoboTwin complement those
near-ceiling tasks.

Figure~\ref{fig:pairedexecutions} shows similar initial approaches but
different final alignment on PushT and goal attainment on Caddy.

\subsection{Which components matter?}
\begin{figure*}[t]
\centering
\begin{minipage}[t]{0.48\textwidth}
\centering
\includegraphics[width=\columnwidth]{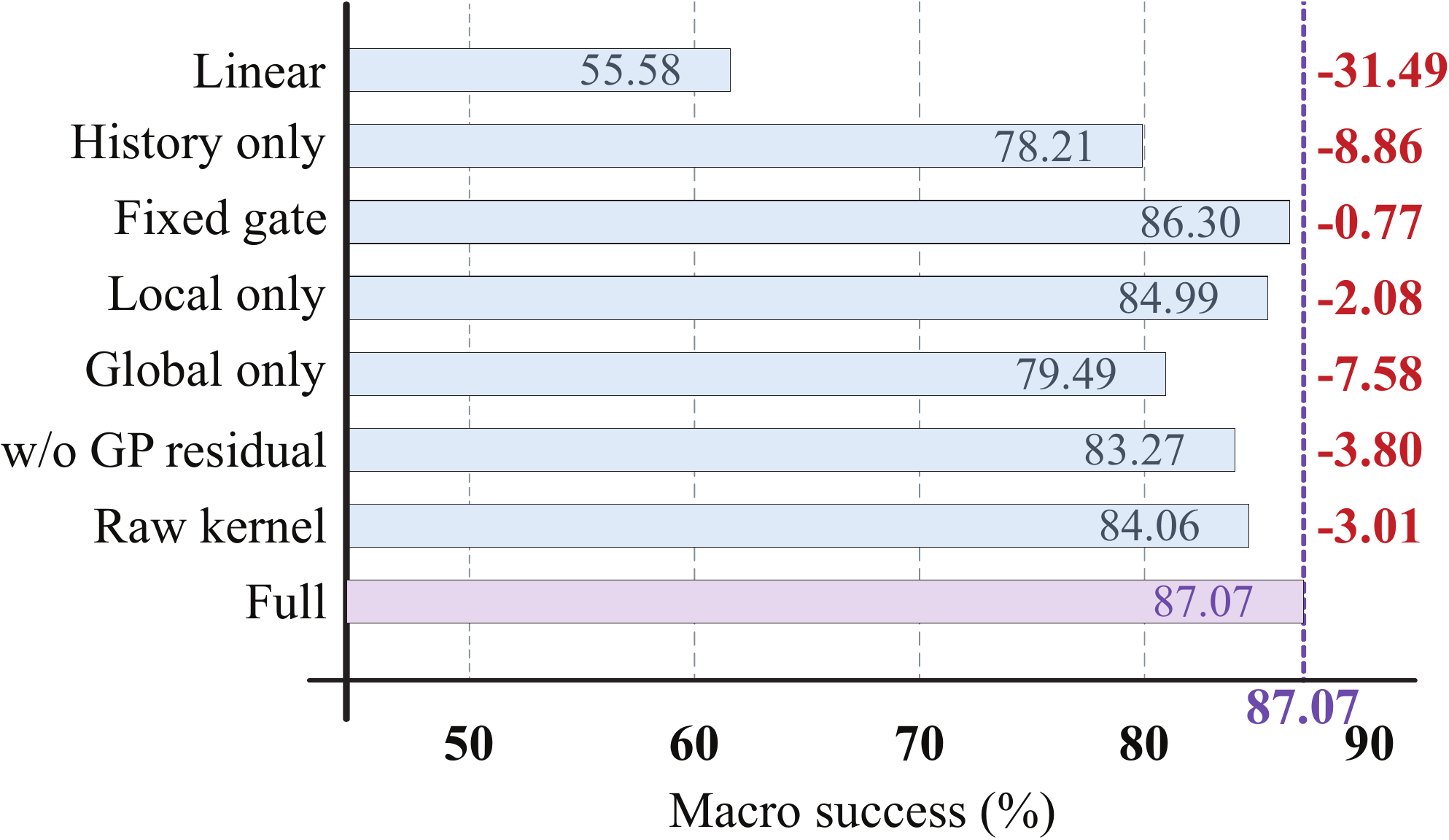}
\par\small (a) Method and component comparisons
\end{minipage}\hfill
\begin{minipage}[t]{0.48\textwidth}
\centering
\includegraphics[width=\columnwidth]{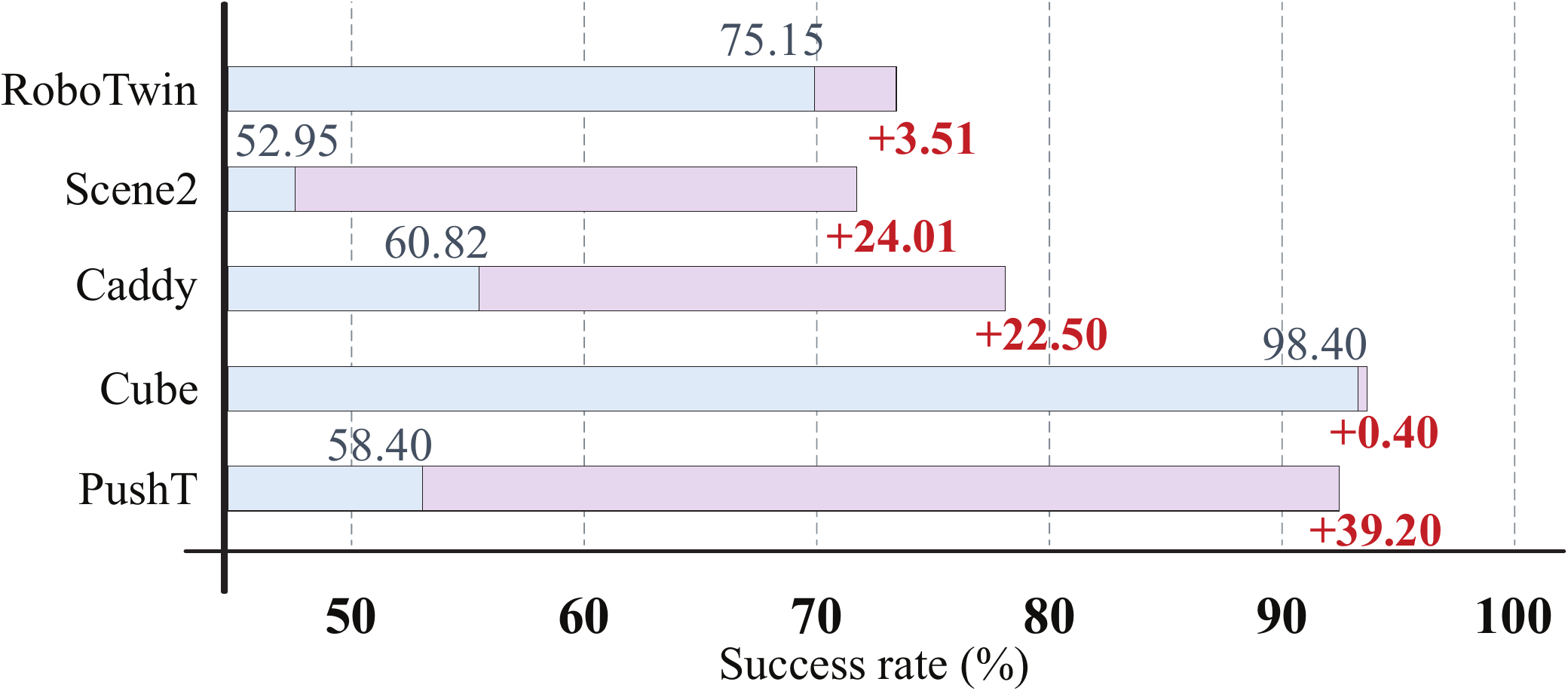}
\par\small (b) Prior execution versus search
\end{minipage}
\caption{\textbf{Contributions of the learned prior and goal-directed search.}
(a) Macro success for prior families and component interventions; caps and
signed annotations show differences from the full model (87.07\%).
(b) Per-task success from posterior-mean execution (blue) and the additional
gain from CEM (light segments). Both use the full neural-plus-GP mean.
Both displayed axes start at 50\%.}
\label{fig:ablation}
\label{fig:search}
\end{figure*}

Figure~\ref{fig:ablation}(a) isolates the contributions of conditioning,
the GP residual, and gated attention. History only removes visual conditioning
from the prior while retaining visual goal evaluation. Its macro success is
8.86 points below LG-DKGP, but it outperforms the full model on Caddy,
Scene2, and RoboTwin: action history supplies strong structure, and visual
conditioning has task-dependent value.

Removing the GP residual while retaining search reduces macro success by
3.80 points, chiefly through Cube's 16.40-point decrease; Scene2 instead
improves by 1.25 points. Independently trained fixed-gate, local-only, and global-only
variants reduce macro success by 0.77, 2.08, and 7.58 points,
respectively, supporting adaptive fusion with a strong role for recent actions.
Replacing learned GP features with raw inputs under the same kernel family
reduces macro success by 3.01 points after separate fitting.

\subsection{History length and planning schedule}
\begin{figure*}[t]
\centering
\begin{minipage}[t]{0.408\textwidth}
\centering
\includegraphics[width=\columnwidth]{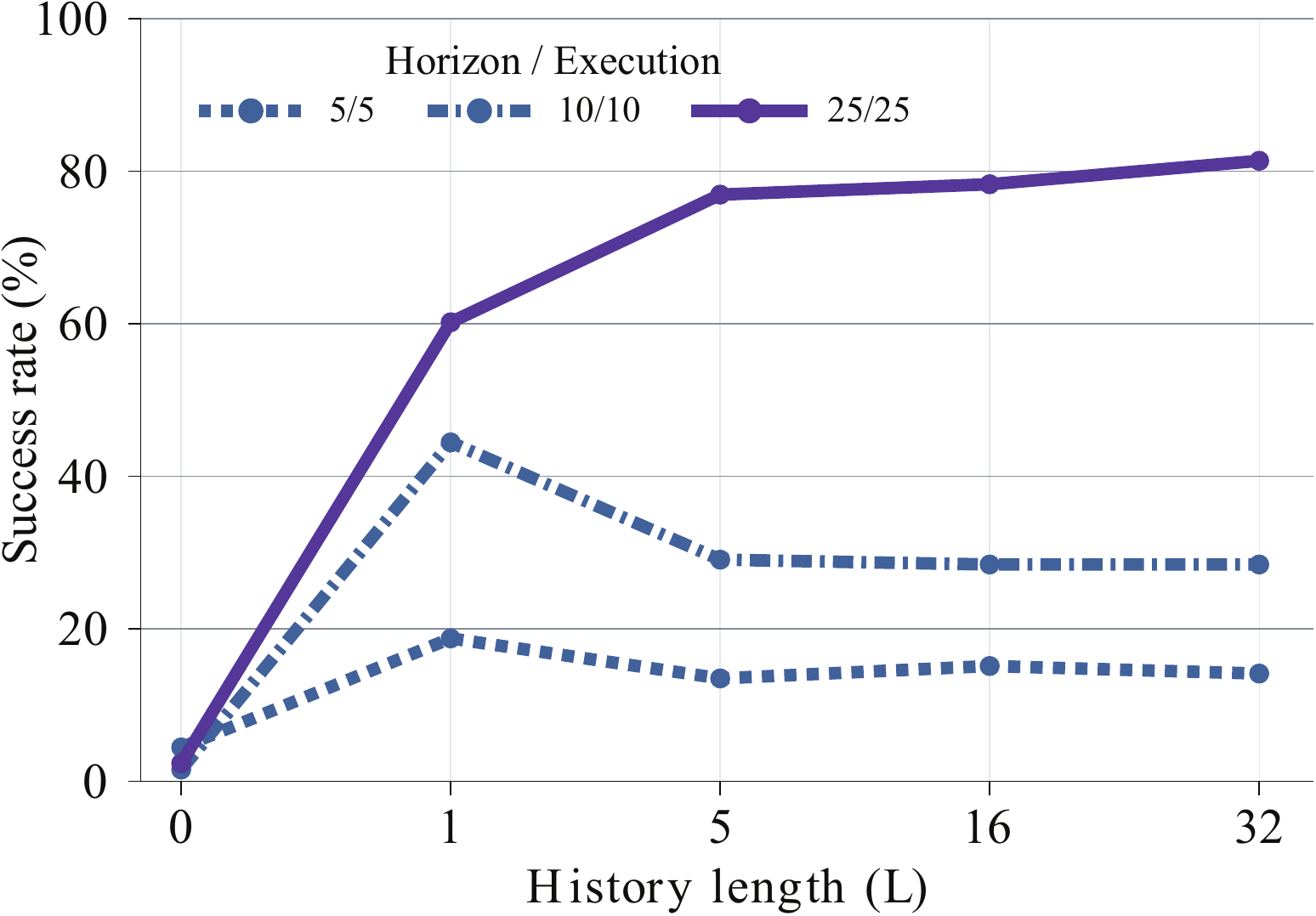}
\par\small (a) History length and planning schedule
\end{minipage}\hspace{0.034\textwidth}%
\begin{minipage}[t]{0.408\textwidth}
\centering
\includegraphics[width=\columnwidth]{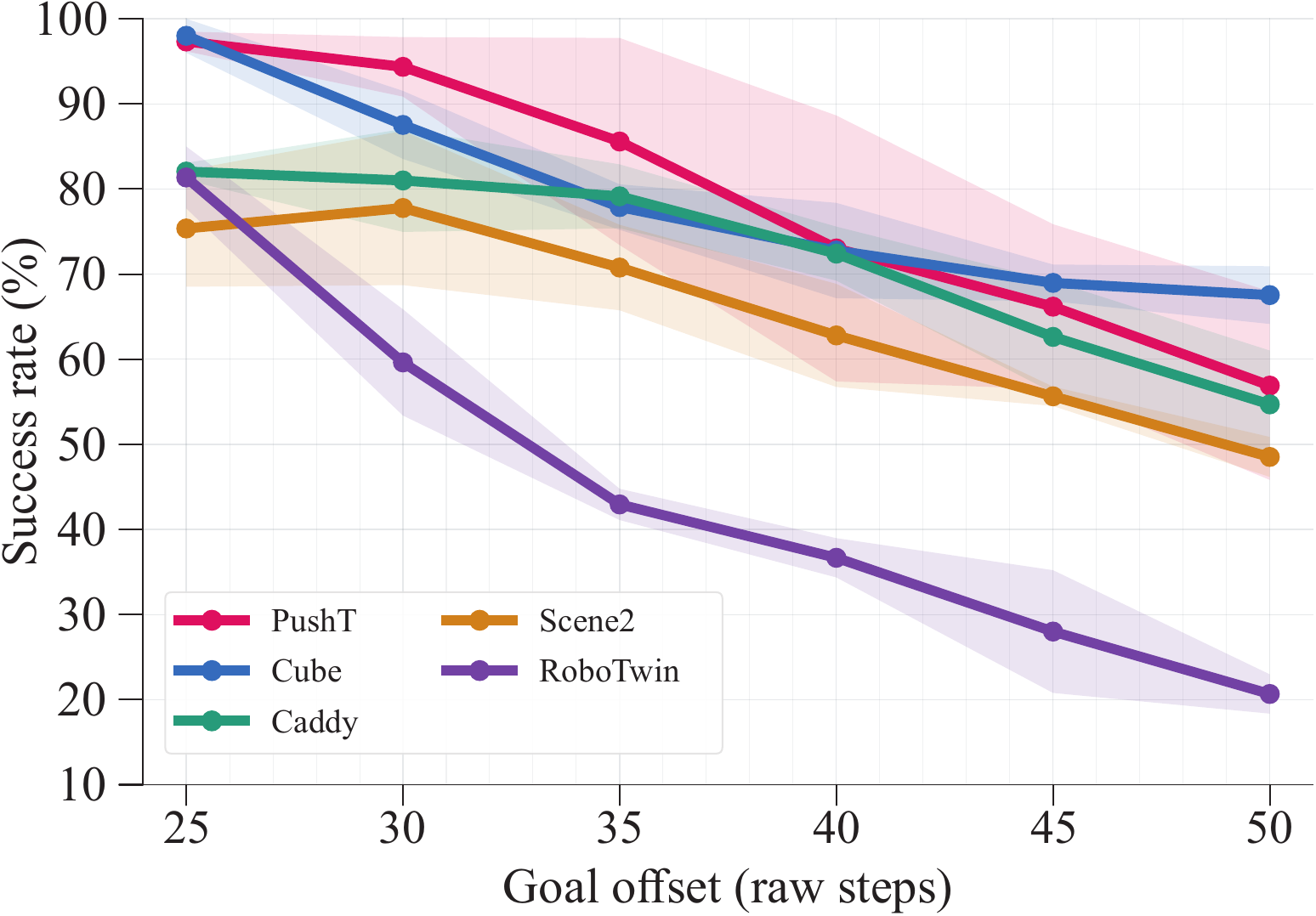}
\par\small (b) Goals beyond the forecast horizon
\end{minipage}
\caption{\textbf{Sensitivity to history, planning schedule, and goal offset.}
Means use seeds 101/202/303. (a) RoboTwin with
$L\in\{0,1,5,16,32\}$, equally spaced on the horizontal axis.
Solid: $H/P=25/25$; distinct dashed lines: $5/5$ and $10/10$.
(b) Five-environment goal-offset study with $H=P=25$, history capped at 32,
and a 50-step budget. Offset 25 reuses existing results for these seeds;
30--50 use fixed execution.}
\label{fig:history}
\label{fig:offset}
\end{figure*}

Figure~\ref{fig:history}(a) uses the complete model, seeds 101/202/303,
and 41/46/46 eligible fragments. History is masked without refitting:
$L=0$ retains only current/goal visual features; $L=1$ includes the latest
action but no adjacent-history difference.

With $H=P=25$, success increases from 2.35\% at $L=0$ to 60.18\% at
$L=1$, 76.92\% at $L=5$, and 81.35\% at $L=32$. With $H=P=10$,
success peaks at 44.43\% for $L=1$, rather than increasing monotonically
with history. The $H=P=5$ schedule reaches at most 18.73\%. Source-prefix
actions initialize the restored scene and causal history before the 50-step
budget. At most two, five, and ten planning calls are allowed, respectively,
jointly varying forecast horizon and replanning frequency.

\subsection{Prior execution versus goal-directed search}
Mean only directly executes the final neural-plus-GP posterior mean. It is
therefore different from \emph{w/o GP residual}, which retains search around
a neural-only center. Figure~\ref{fig:search}(b) shows gains from 58.40\% to
97.60\% on PushT, 60.82\% to 83.32\% on Caddy, 52.95\% to 76.96\%
on Scene2, and 75.15\% to 78.66\% on RoboTwin. Cube is already at 98.40\%
without search and improves by 0.40 points. Plausible continuations and
model-based goal selection play complementary roles.

\subsection{How should search be initialized?}

Table~\ref{tab:initialization} compares four initializations with the same
fitted prior, world model, queries, and planning budget. Expert isotropic
scaling improves macro success from 74.92\% with fixed scaling to 87.07\%.
RoboTwin shows the largest gain, from 18.88\% to 78.66\%, using a much
smaller proposal scale. PushT is unchanged, while Cube slightly favors the
fixed scale. Diagonal calibration and increment bounds trail isotropic
scaling overall, supporting expert-derived exploration without hard
bounds on action changes.

\begin{table}[t]
\centering
\caption{\textbf{Search initialization with the same learned prior (\%).}
Five-seed means. Fixed uses $0.3^2\mathbf I$; Diag. uses expert-derived
coordinatewise scales after subtracting the mean continuation's increment
energy; +Bounds additionally applies action bounds and a
training-derived increment budget. Ours uses the expert-derived isotropic
scale without these bounds.}
\label{tab:initialization}
\setlength{\tabcolsep}{3pt}\footnotesize
\begin{tabular*}{\columnwidth}{@{\extracolsep{\fill}}lrrrr@{}}
\toprule
Task & Fixed & Diag. & +Bounds & \textbf{LG-DKGP}\\
\midrule
\InitializationRows
\bottomrule
\end{tabular*}
\end{table}

\subsection{Goals beyond the forecast horizon}
Figure~\ref{fig:offset}(b) pursues goals 25--50 demonstration steps ahead while
keeping $H=P=25$. The 75 seed-level units at offsets 30--50 contain 3,358
eligible evaluations. Between offsets 30 and 50, success decreases from
94.35\% to 56.90\% on PushT, 87.53\% to 67.53\% on Cube, 81.00\%
to 54.69\% on Caddy, 77.77\% to 48.52\% on Scene2, and 59.63\% to
20.67\% on RoboTwin. Offset counts demonstration steps, with eligibility
recomputed for each goal query; the offset-25 reference retains its original
cohort. The curves reveal task-dependent limits beyond the learned forecast
horizon.

\subsection{Do generated actions better match expert structure?}
\begin{figure*}[t]
\centering
\includegraphics[width=\textwidth]{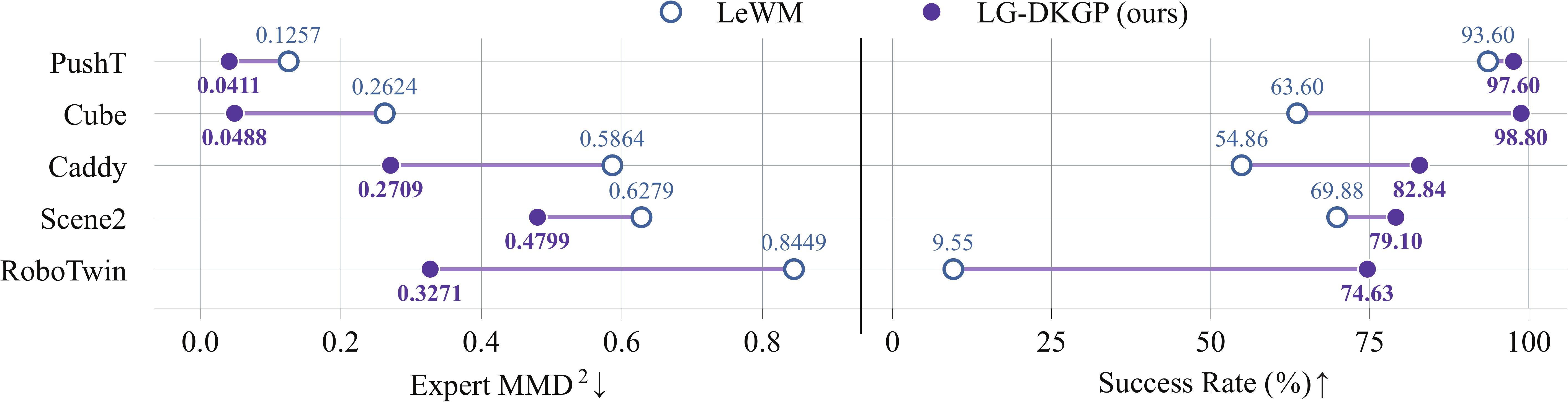}
\caption{\textbf{Expert-relative action structure and control.} Hollow and
filled markers denote LeWM and LG-DKGP. Left: expert transition-feature
MMD$^2$ (lower is better). Right: closed-loop success (higher is better).
All eligible plans enter the structure comparison, including failures.}
\label{fig:alignment}
\end{figure*}

Figure~\ref{fig:alignment} compares LG-DKGP and original LeWM on matched
scene/goal queries using their frozen world models, $H=P=25$, a 50-step
budget, and five seeds. The first complete optimized plan is recorded before
success-dependent termination and compared with its expert continuation.

For standardized continuous actions, define
$\Delta_\tau\widetilde{\mathbf a}_i^c=
\widetilde{\mathbf a}_{i+\tau}^c-\widetilde{\mathbf a}_i^c$ and transition features
\begin{equation}
 \mathbf v_i=[\Delta_1\widetilde{\mathbf a}_i^c;
 \Delta_1\widetilde{\mathbf a}_{i+1}^c;
 \Delta_2\widetilde{\mathbf a}_i^c;
 \Delta_5\widetilde{\mathbf a}_i^c],\quad i=0,\ldots,19.
\end{equation}
Gaussian-kernel MMD$^2$~\cite{gretton2012kernel} compares generated and expert
feature sets, with bandwidth calibrated on expert training features. It
retains signed changes and multi-step relations, rather than only variation
magnitudes. A complementary profile error compares squared action differences
at lags $1,2,3,5,10$ against the expert profile.
Gripper transitions are handled separately.

LG-DKGP reduces MMD$^2$ and profile error and increases mean success in all
five aligned-query sets. Mean MMD$^2$ reductions range from 0.0846 to
0.5178; all five source-episode-cluster bootstrap intervals exclude zero.
Success gains range from 4.00 to 65.08 points. The Caddy, Scene2, and RoboTwin
rates differ from Table~\ref{tab:main} because this is a separate matched-query
study. The paired structure and control improvements support the use of
expert-informed proposals.

\begin{figure}[t]
\centering
\includegraphics[width=\columnwidth]{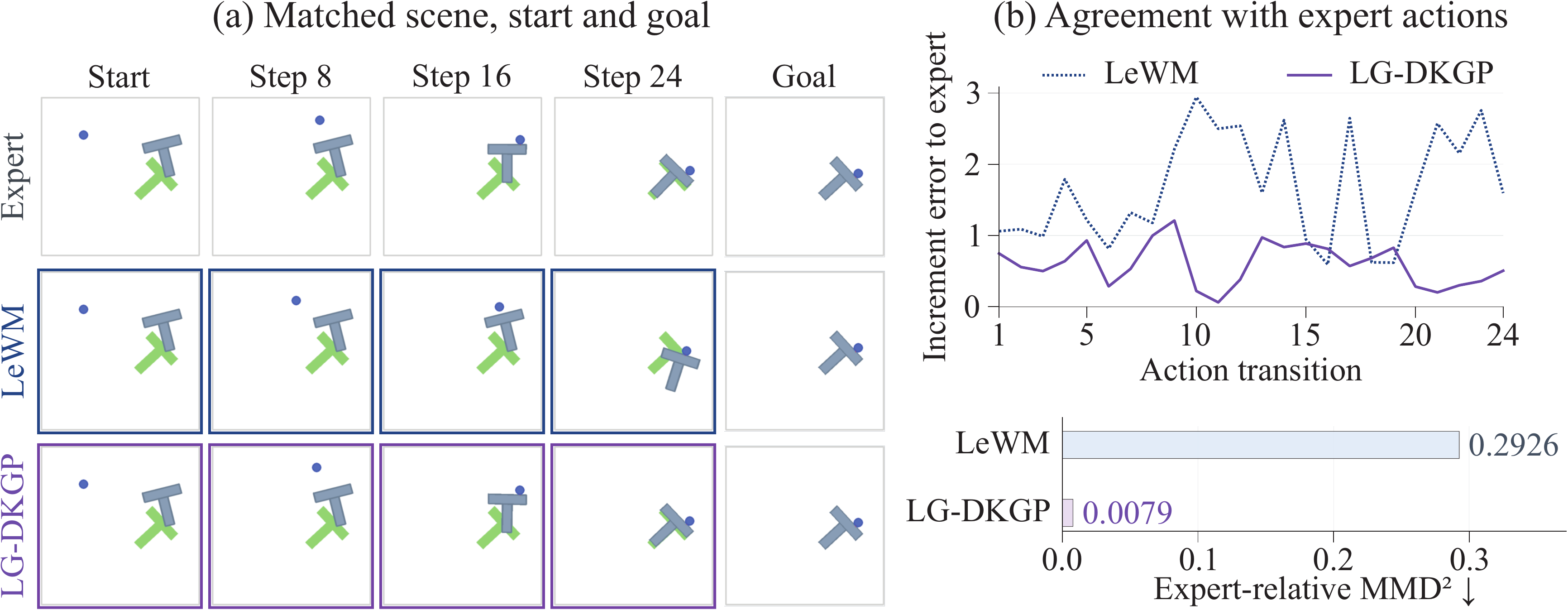}
\caption{\textbf{Expert-relative structure on PushT.}
Left: expert, LeWM, and LG-DKGP executions at matched steps from the same
start and goal. Right: expert-relative adjacent-increment error and
MMD$^2$ for the first optimized 25-action plans (LeWM: 0.2926;
LG-DKGP: 0.0079).}
\label{fig:expertcase}
\end{figure}

In Fig.~\ref{fig:expertcase}, LG-DKGP follows the expert's object poses more
closely, with generally smaller adjacent-increment errors and lower MMD$^2$
alongside successful goal attainment.

\subsection{Is expert alignment consistent across queries?}
\begin{table}[t]
\centering
\caption{\textbf{Paired expert-relative structure.} Closer counts queries where
LG-DKGP has lower MMD$^2$ than LeWM. $\Delta$ is LeWM minus LG-DKGP;
intervals use 1,000 source-episode-cluster bootstrap resamples. Query counts
include both successful and failed control episodes.}
\label{tab:pairedstructure}
\setlength{\tabcolsep}{2pt}\footnotesize
\begin{tabular*}{\columnwidth}{@{\extracolsep{\fill}}lrrrr@{}}
\toprule
Task & Closer & (\%) & $\Delta$ & 95\% interval\\
\midrule
\PairedStructureRows
\bottomrule
\end{tabular*}
\end{table}

Table~\ref{tab:pairedstructure} examines expert alignment for individual
scene/goal pairs. LG-DKGP is closer to the expert continuation
on 83.6\% of PushT queries, 90.0\% of Cube queries, 99.5\% of Caddy queries,
98.2\% of Scene2 queries, and all 219 RoboTwin queries. Repeated source
episodes remain within the same bootstrap cluster.

We evaluate half and twice the expert-calibrated bandwidth. LG-DKGP retains lower task-mean MMD$^2$ than LeWM in every
environment at both settings.

\paragraph{Metric stress tests.}
Two offline references test the metric: repeating the expert continuation's
first action for 25 steps and randomly permuting its 25 actions. The constant reference still
obtains lower MMD$^2$ than LG-DKGP on Caddy, Scene2, and RoboTwin,
despite discarding the expert's temporal evolution.
This exposes a limitation of finite-sample transition-feature MMD as a
standalone action-quality measure. Moreover, lower discrepancy does not
consistently separate successful and failed plans for each planner.
We therefore evaluate transition discrepancy together with expert variation
profiles and closed-loop success.

\section{Discussion and Limitations}
\label{sec:limitations}
LG-DKGP turns expert action continuity into an inductive bias for planning. Gated history
and the neural/GP mean condition proposed continuations on past actions;
expert-derived exploration and frozen-model goal evaluation adapt these
continuations toward the task goal.

\paragraph{Task-dependent conditioning.}
History-only proposals outperform visual-conditioned proposals on three
tasks, reinforcing this principle.
At smaller $H,P$, the appropriate history length and the relative
contributions of action history and observations to success remain open.

\paragraph{Forecast support.}
Beyond the 25-action forecast support, success declines with goal distance.
This exposes a broader world-model planning bottleneck~\cite{maes2026lewm}
that limits wider application. Future work should sustain short-horizon
success rates on long-horizon tasks while preserving expert-like action-trajectory
continuity.

{\small
\bibliographystyle{ieeenat_fullname}
\bibliography{references}
}
\end{document}